\documentclass[conference]{IEEEtran}

\usepackage{cite}
\usepackage{balance}
\usepackage{amsmath,amssymb,amsfonts}
\usepackage{algorithmic}
\usepackage{graphicx}
\usepackage{textcomp}
\usepackage{xcolor}
\def\BibTeX{{\rm B\kern-.05em{\sc i\kern-.025em b}\kern-.08em
    T\kern-.1667em\lower.7ex\hbox{E}\kern-.125emX}}

\usepackage{float}
\usepackage{hyperref}
\usepackage[inkscapelatex=false]{svg}

\begin{document}

\title{Multivariate Time Series Clustering for\\
Environmental State Characterization of\\
Ground-Based Gravitational-Wave Detectors}

\author{\IEEEauthorblockN{Rutuja Gurav}
\IEEEauthorblockA{\textit{Computer Science \& Engineering}\\
\textit{University of California, Riverside}\\
Riverside, CA, USA\\
rutuja.gurav@email.ucr.edu}
\and
\IEEEauthorblockN{Isaac Kelly}
\IEEEauthorblockA{\textit{Physics}\\
\textit{University of Dallas}\\
Dallas, TX, USA\\
ikelly@udallas.edu}
\and
\IEEEauthorblockN{Pooyan Goodarzi}
\IEEEauthorblockA{\textit{Physics \& Astronomy}\\
\textit{University of California, Riverside}\\
Riverside, CA, USA\\
pooyan.goodarzi@email.ucr.edu}
\and
\IEEEauthorblockN{Anamaria Effler}
\IEEEauthorblockA{\textit{LIGO Laboratory}\\
\textit{LIGO Livingston Observatory}\\
Livingston, LA, USA\\
aeffler@ligo-la.caltech.edu}
\and
\IEEEauthorblockN{Barry Barish}
\IEEEauthorblockA{\textit{Physics \& Astronomy}\\
\textit{University of California, Riverside}\\
Riverside, CA, USA\\
barry.barish@ucr.edu}
\and
\IEEEauthorblockN{Evangelos E. Papalexakis}
\IEEEauthorblockA{\textit{ Computer Science \& Engineering}\\
\textit{University of California, Riverside}\\
Riverside, CA, USA\\
epapalex@cs.ucr.edu}
\and
\IEEEauthorblockN{Jonathan W. Richardson}
\IEEEauthorblockA{\textit{Physics \& Astronomy}\\
\textit{University of California, Riverside}\\
Riverside, CA, USA\\
jonathan.richardson@ucr.edu}
}

\maketitle

\thispagestyle{plain}
\pagestyle{plain}

\begin{abstract}
Gravitational-wave observatories like LIGO are large-scale, terrestrial instruments housed in infrastructure that spans a multi-kilometer geographic area and which must be actively controlled to maintain operational stability for long observation periods. Despite exquisite seismic isolation, they remain susceptible to seismic noise and other terrestrial disturbances that can couple undesirable vibrations into the instrumental infrastructure, potentially leading to control instabilities or noise artifacts in the detector output. It is, therefore, critical to characterize the seismic state of these observatories to identify a set of temporal patterns that can inform the detector operators in day-to-day monitoring and diagnostics. On a day-to-day basis, the operators monitor several seismically relevant data streams to diagnose operational instabilities and sources of noise using some simple empirically-determined thresholds. It can be untenable for a human operator to monitor multiple data streams in this manual fashion and thus a distillation of these data-streams into a more human-friendly format is sought. In this paper, we present an end-to-end machine learning pipeline for features-based multivariate time series clustering to achieve this goal and to provide actionable insights to the detector operators by correlating found clusters with events of interest in the detector.
\end{abstract}


\section{Introduction}
\label{intro}

In the last nine years, the Laser Interferometer Gravitational-wave Observatory (LIGO)~\cite{LIGOScientific:2014pky} and the European Virgo observatory~\cite{VIRGO:2014yos} have established gravitational waves as a new observational probe of the universe. Gravitational waves, disturbances in the geometry of spacetime generated by the acceleration of matter, are a longstanding prediction of general relativity. The first direct detection of gravitational waves from two merging black holes in 2015 by LIGO~\cite{Abbott:2016} opened a new window on the universe. A subsequent observation of a binary neutron star merger with gravitational waves~\cite{Abbott:2017a} and every band of the electromagnetic spectrum~\cite{Abbott:2017b, Abbott:2017c} launched a new \textit{multi-messenger} era of astronomy. The current generation of detectors have now observed a wide variety of merger events involving black holes and neutron stars~\cite{Abbott:2021, GWTC-3:2021vkt, Olsen:2022pin, Nitz:2021zwj}. The insights provided by these gravitational-wave signals are being used to address long-standing questions in astrophysics and fundamental physics.

\begin{figure}[tbp]
    \centering
    \includegraphics[width=\columnwidth]{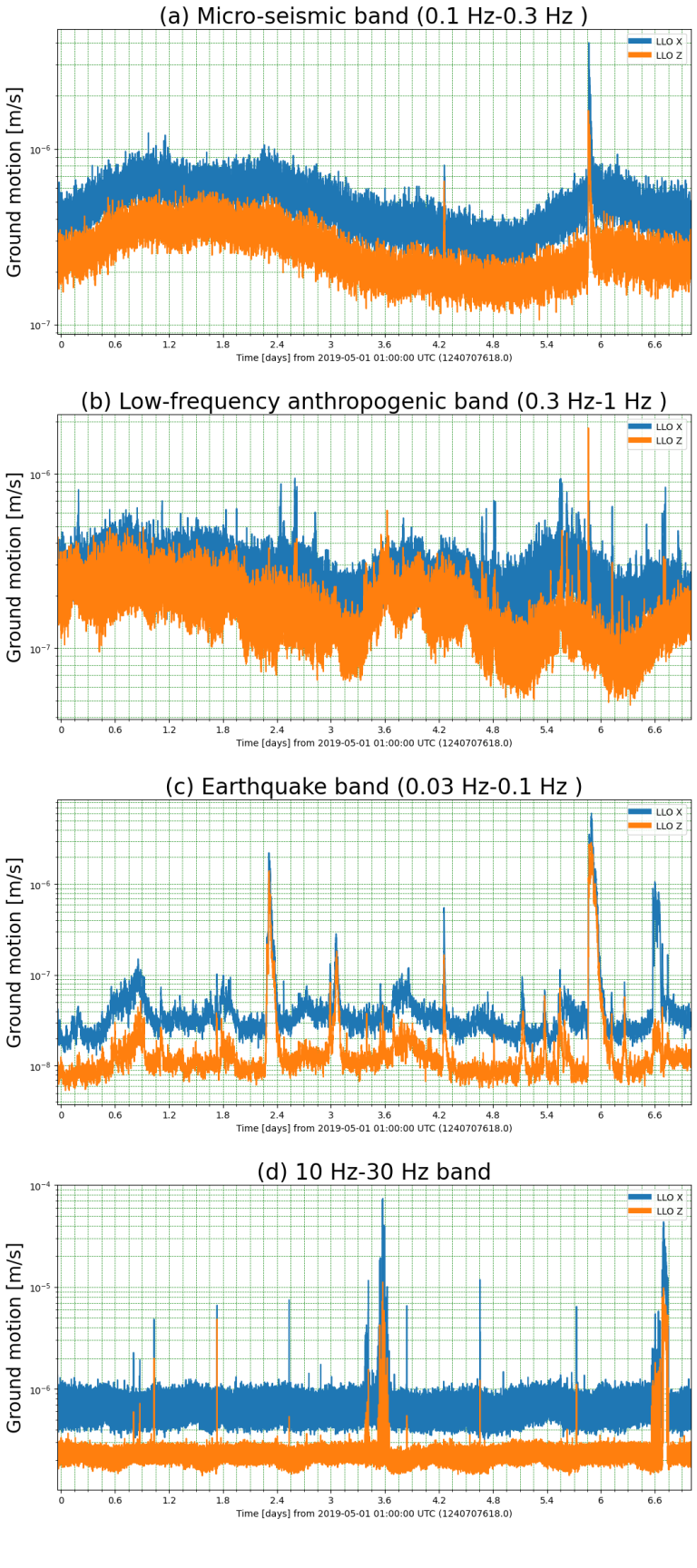}
    \caption{A one-week band-limited sample of root-mean-square (RMS) ground motion data recorded by seismometers at the LIGO Livingston site. Each frequency band is associated with a set of different physical causes. (a) Micro-seismic frequency band (0.1-0.3~Hz) is mostly sensitive to ground motion caused by oceanic waves and has a characteristic time scale of multiple days~\cite{giaime2003feedforward}. (b) Low-frequency anthropogenic band (0.3-1~Hz) is correlated to various human related activities and the tides. (c) Earthquake band (0.03-0.1~Hz) captures ground motions mostly due to earthquakes and wind. (d) The 10-30~Hz frequency band is sensitive to ground motion due to mechanical vibrations of equipment at the LIGO sites, such as the HVAC system~\cite{Nguyen:2021}.}
    \label{fig:sample_time_series}
\end{figure}

Gravitational-wave detectors like LIGO are large-scale, terrestrial instruments housed in an infrastructure which sprawls a large geographic area. The two LIGO detectors, located in Hanford, WA and Livingston, LA, each consist of a 4-km-long Michelson laser interferometer whose sensitivity is further enhanced using multiple internal laser cavities~\cite{Martynov:2016}. The length and angular degrees of the individual laser cavities, as well as the interferometer as a whole, must be sensed and actively controlled in order to maintain operational stability for long observation periods. Despite exquisite seismic isolation of the detectors' optics, they remain susceptible to seismic noise and other terrestrial disturbances that can couple undesirable vibrations into the instruments' infrastructure. Figure~\ref{fig:sample_time_series} shows an example of some of the environmental disturbances regularly recorded by sensors at the LIGO sites.

By introducing physical motions between the interferometers' optics, environmental disturbances limit LIGO's sensitivity to gravitational waves at the lower end of its sensitive band, below 20~Hz. Environmental noise also poses a serious challenge to the operational stability and data quality of the detectors. For example, environmental conditions may account for many detector \textit{glitches}, nonstationary noise bursts of largely unknown origin, that severely degrade the detector sensitivity during their duration~\cite{Blackburn:2008, Cabero:2019, LIGO:2021ppb}. Glitches contaminate the astrophysical data streams, confusing the gravitational-wave search pipelines and hindering the timely issuance of real-time alerts for electromagnetic follow-up observations. Elevated environmental noise is also known to cause \textit{lock losses}, control failures occurring when a disturbance causes 

\noindent
the laser cavities to become driven too far from their resonant operating points~\cite{Rollins:2017}. Because lock re-acquisition typically requires 30~minutes to complete, lock losses limit the duty cycle for multi-detector observations, which are essential for precise sky localization of gravitational-wave events.

As a result, LIGO is extremely interested in understanding the emergence and effects of external disturbances and monitoring their behavior over time. For example, periods of increased glitch rates have been anecdotally associated with elevations of certain types of environmental noise. In fact, internally, detector commissioners already have heuristic means for manually monitoring such behavior. The motivation of this work is a result of direct collaboration with LIGO in order to automate, and extend, this endeavor. Our objective is to distill the information from a large number of heterogeneous environmental sensors, distributed across the 4-km LIGO sites, into a single environment ``state'' word that can be continuously recorded and tracked over time. Beyond automation, this tracking has the potential to reveal new, previously unrecognized associations between specific environmental conditions and detector anomalies (e.g., periods of increased glitch rates or controls instabilities). Such associations can provide actionable insight into the physical nature of the anomalies, directly guiding detector commissioning.

In this paper, we present an end-to-end multivariate time series analysis pipeline built to characterize the environmental state of ground-based gravitational-wave observatories like LIGO. The task at hand is to \textit{identify known seismic and other environmental phenomena (e.g., earthquakes, anthropogenic activity) in measurements made by the network of sensors deployed across the detector sites and assign an interpretable label to each point in time identifying combinations of phenomena and the location(s) where they are active}. The major aspects of this work include:
\begin{itemize}
    \item {\bf End-to-end data science pipeline}: A major contribution of this work is identifying what data sources to monitor and collect and how to translate the task, defined by a real LIGO need, to a data science pipeline. Given our active collaboration with LIGO, this pipeline has the potential to be deployed at the sites as a powerful diagnostic tool.
    \item {\bf Dataset}: We accompany our work with a public release of a relevant dataset that can foster further research in this direction. The dataset includes all of the LIGO time series data used in the development of our pipeline, hosted and provided by the Gravitational Wave Open Science Center (GWOSC)~\cite{KAGRA:2023pio, O3-data-trend}.\footnote{Data available at: \url{https://gwosc.org/O3/trend/}}
\end{itemize}

The rest of this paper is organized as follows. In Section~\ref{background} we provide a brief background on state-of-the-art gravitational-wave detectors, followed by a review of the existing work related to environmental state characterization. Section~\ref{method} presents our proposed data science pipeline. We then present results on a real LIGO dataset in Section~\ref{results} to demonstrate the effectiveness of this pipeline. Finally, we conclude this paper in Section~\ref{conclusion}.
\section{Background and Motivation}
\label{background}

\subsection{Gravitational-Wave Detectors}
Gravitational-wave detectors are multi-kilometer-scale instruments ranking among the largest and most complex scientific facilities in the world. In addition to the main channel sensitive to spacetime strain, where gravitational-wave signals are observed, each LIGO detector has over 100,000~auxiliary channels which monitor the operation of each subsystem and the seismic, acoustic, and electromagnetic environment. LIGO's physical environmental monitoring~(PEM) system~\cite{Effler:2015, Nguyen:2021, acernese2022virgo} consists of a distributed network of accelerometers, seismometers, microphones, magnetometers, power-mains voltage monitors, radio-frequency receivers, cosmic-ray detectors, and wind, temperature, and humidity sensors. This wealth of data presents a unique and largely untapped opportunity: \emph{Can we leverage these vast amounts of data to improve our understanding of the relationship between changing environmental conditions and their impact on the detector's performance?}

\subsection{Relation to Previous Work}
In the context of environmental state characterization, a prior pipeline has been developed within LIGO to provide real-time seismic predictions to the interferometer operators. Seismon~\cite{Coughlin:2017} is an earthquake early-warning system deployed at both LIGO detector sites. It uses near real-time earthquake alerts provided by the U.S. Geological Survey (USGS) and the National Oceanic and Atmospheric Administration (NOAA) to estimate the time of arrival and amplitude of the surface waves of earthquakes from around the globe at each detector site. Based on the predicted amplitude and direction of the surface waves, Seismon provides a machine-learning-based estimate of the probability that the incoming seismic event will cause a lock loss.

Seismon has enhanced the operational stability and up-time of the LIGO detectors by providing operators with the information necessary for on-site decision making to transition the detector to a more earthquake-robust operational mode. ``Earthquake mode'' is an alternative controls strategy which allows the detectors' length servos to handle larger seismic disturbances, but at the expense of increased instrumental noise (reduced astrophysical sensitivity). The work presented here plays a highly complementary role to Seismon. Rather than making forward-looking predictions for real-time decision making, our pipeline is designed to characterize the \textit{current} environmental state and can potentially include many other forms of disturbances, beyond seismic phenomena, as well.

Furthermore, in a broader context of state characterization, various detector characterization methods are used to generate data quality products from both the main strain channel and the auxiliary channels. These products take the form of bit-vector flags that indicate time segments of specific quality within each interferometer~\cite{LIGO:2021ppb}. For instance, there are "observing mode" flags at both sites generated by detector operators, indicating time segments in which the interferometer is locked and is reliable for astrophysical data inference. In this study, we only utilized the segments flagged in this manner, defining the detector's status as being in observing mode.

\subsection{Related Time Series Work}
There exist a number of relevant works in the time series mining and learning literature that have tackled problems that are similar to the one at hand. Matsubara~\textit{et al.}~\cite{matsubara2014autoplait} has developed AutoPlait, a co-evolving time series mining tool that can identify a general set of patterns among a collection of time series that are related. There is even earlier work by Papadimitriou~\textit{et al.}~\cite{papadimitriou2005streaming} that attempts to do so, with the additional constraint that the data is seen as a stream, which introduces computational challenges.

Beyond the general-purpose identification of patterns in multiple time series, there have been various problem definitions and associated solutions which can fit our scenario and are also highly related to each other. These include regime shifts in multivariate time series~\cite{matsubara2016regime}, change detection~\cite{hooi2019branch}, and multivariate time series segmentation~\cite{gharghabi2019domain}. Broadly, these techniques all seek to identify periods of correlated behavior across a collection of time series and points in time where those periods change from one category to another. Additionally, works that detect anomalies in multivariate time series~\cite{deng2021graph,paltenghi2020time,tafazoli2023matrix, laguarta2024detection}, while not directly addressing the problem definition at hand, computationally require a similar approach. Here, the interest is in identifying irregular patterns that far exceed the ``normal'' behavior, which the rest of the related works seek to characterize.
\section{Proposed Pipeline}
\label{method}
In this section, we describe our pipeline consisting of three modules:
\begin{enumerate}
    \item Dataset Creation
    \item Modeling
    \item Downstream Evaluation
\end{enumerate}
Our code-base is available at UC Riverside's git repository.~\footnote{\url{https://git.ligo.org/uc_riverside/state-characterization}}

\subsection{Dataset Creation}
\label{subsec:dataset_creation}

\begin{figure}[!ht]
    \centering
    \includegraphics[width=\linewidth, keepaspectratio]{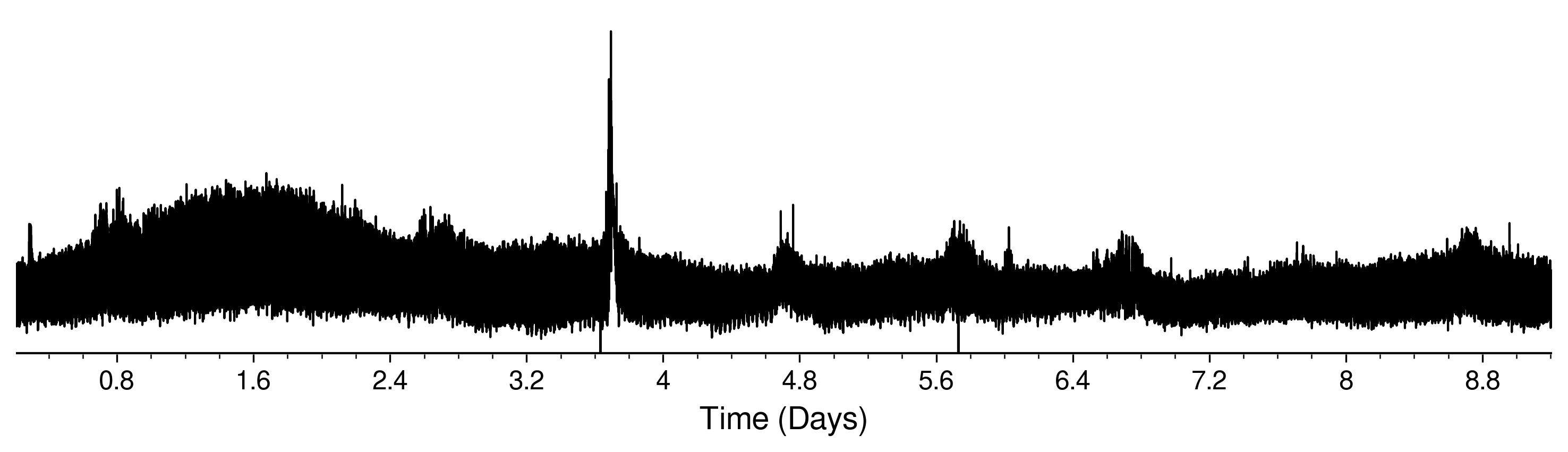}
    \includegraphics[width=\linewidth, keepaspectratio]{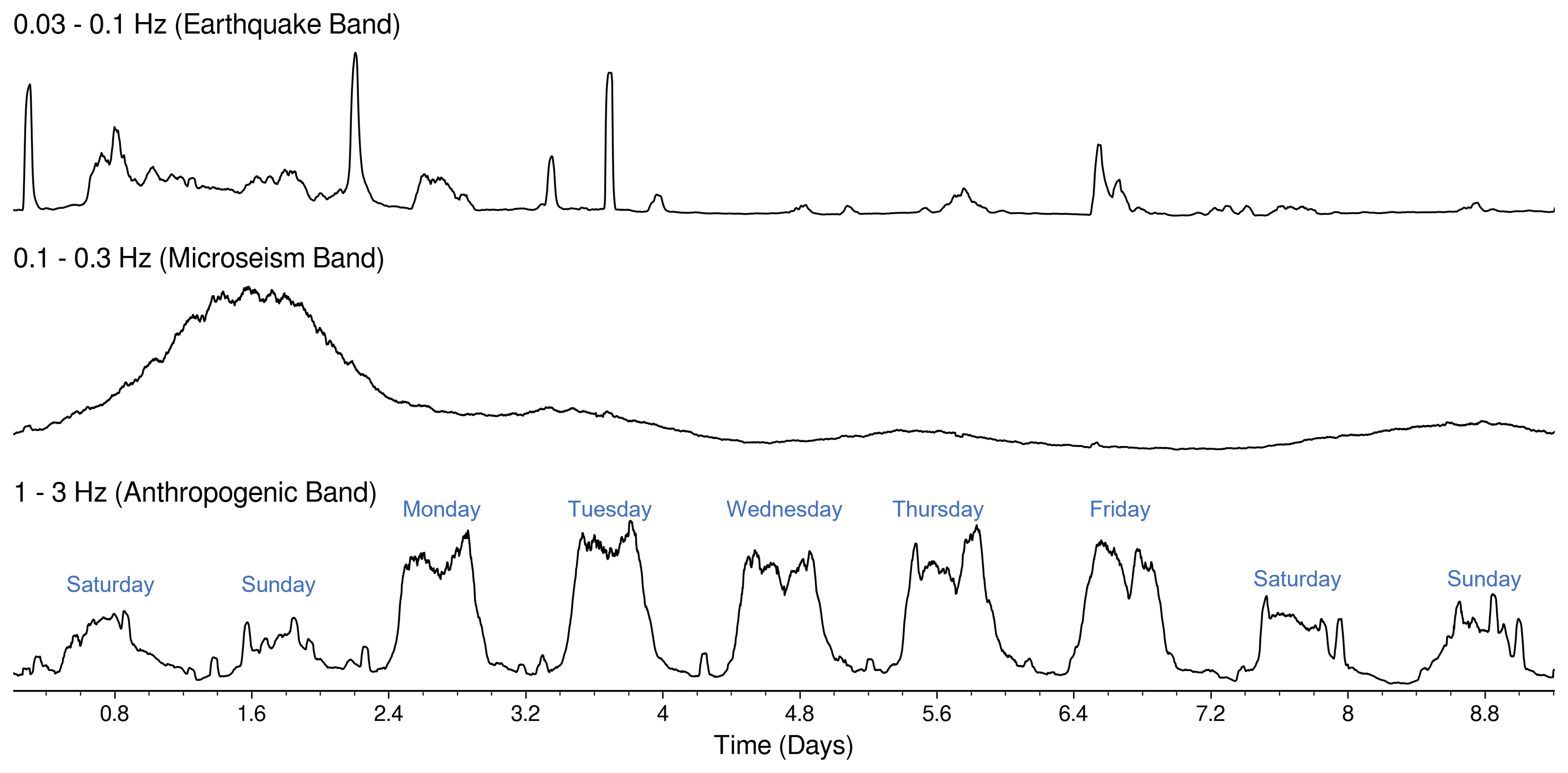}
    \caption{\textit{Top:} Example time series data from one of the numerous seismometers deployed across the LIGO sites. \textit{Bottom:} The same signal bandpass-filtered to select three physically-motivated frequency bands corresponding to known seismic phenomenon.}
    \label{fig:data_and_postproc}
\end{figure}

\paragraph{Data}
There are numerous seismometers deployed across the detector sites to monitor seismic activity. Figure~\ref{fig:data_and_postproc}~(top) shows a readout from one of these sensors. There are known seismic phenomena that manifest in certain frequency ranges, Figure~\ref{fig:data_and_postproc}~(bottom) shows three time series which are obtained by bandpass filtering the seismometer readout to isolate the following physically-motivated frequency ranges.

\begin{figure*}[!ht]
    \centering
    \includegraphics[width=\linewidth]{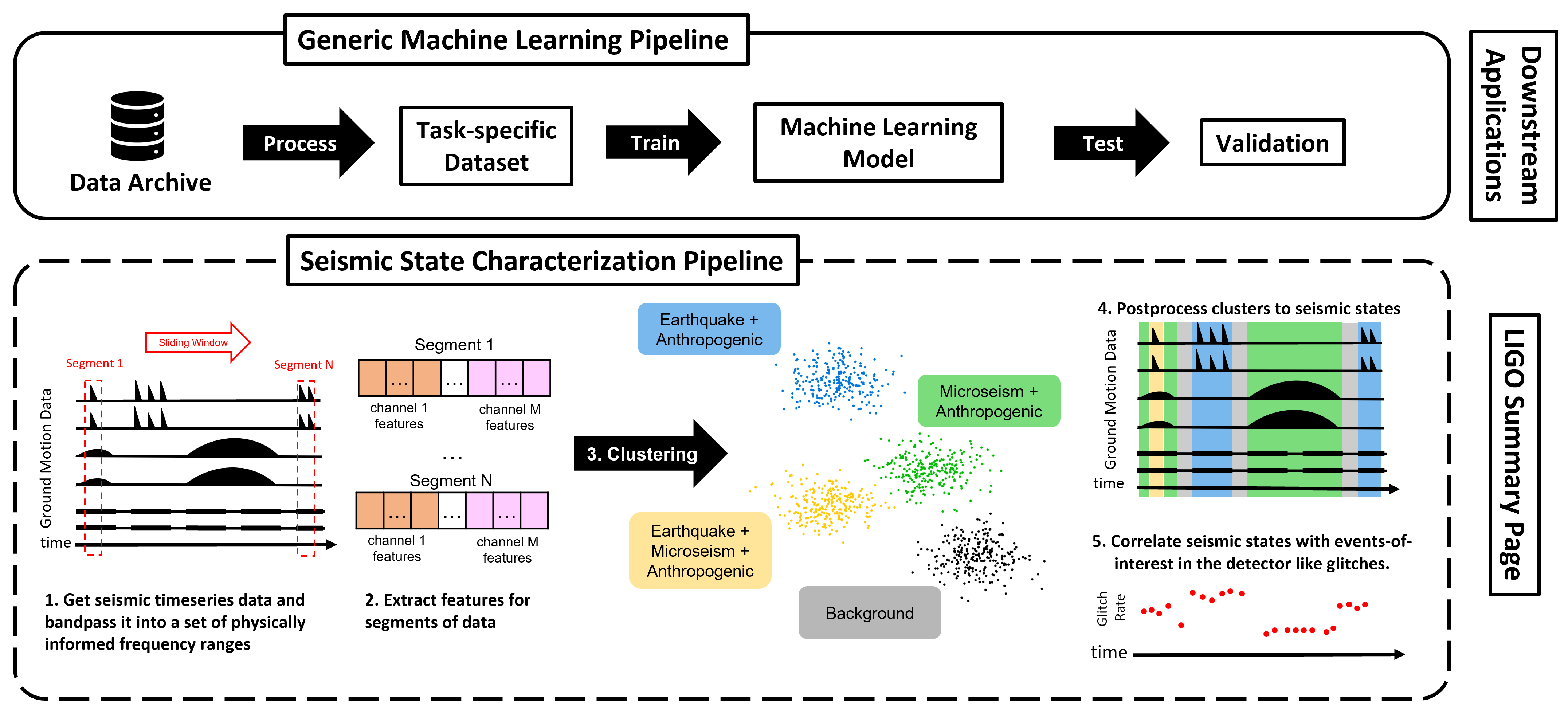}
    \caption{Step-by-step workflow of our proposed pipeline. Each step is described in detail in the text of \S\ref{subsec:modeling}.}
    \label{fig:pipeline}
\end{figure*}

\begin{enumerate}
    \item \textbf{0.03-0.1 Hz Earthquake band:} This frequency band is sensitive to ground motion due to earthquakes.
    \item \textbf{0.1-0.3 Hz Microseism band:} This frequency band is sensitive to ground motion due to ocean waves beating against the shore, dominated by the Pacific Ocean for the LIGO~Hanford detector and the Atlantic Ocean and Gulf of Mexico for the LIGO~Livingston detector.
    \item \textbf{1-3 Hz Anthropogenic band:} This frequency band is sensitive to ground motion due to daily human activity, such as heavy traffic on nearby roads, passing trains, and logging or construction close to a LIGO site.
\end{enumerate}

Although LIGO data channels have various sample rates depending on the physical quantity they measure and the specific sensors used for measurement, in this study, we used the band-limited RMS-averaged (BLRMS) second-trend data from each channel. Therefore, each data point is generated by calculating the root mean square of one-second segments of the recorded data. 

In the current analysis, we used the data from a one month period during the O3b run. The utilized sensors are five tri-axial seismometers (STS) located at different positions of the LIGO Livingston's Internal Seismic Isolation subsystem (L1-ISI), namely at ETMX, ETMY, ITMX, ITMY, and HAM5~\cite{LIGOScientific:2014pky}. Each sensor has three orthogonal axes and is band-limited to six physically motivated frequency bands. The total data volume for the 90 second-trend L1-ISI-STS-BLRMS channels used in this analysis amounts to 20 GiB.

\subsection{Modeling}
\label{subsec:modeling}
In our approach, given that (1) we require near real-time computations that can provide immediate insights to the LIGO operators and (2) we are expecting domain experts to deploy and fine-tune our tool, we would like to start from a simple approach with a minimal number of hyperparameters. As a result, even though (as we outlined in the related work above) there exist a number of off-the-shelf methods that could be adapted for the purposes of this work, we instead opt for a light-weight clustering-based approach which can run fast and requires the definition of only two hyperparameters: the duration of window segments we are considering and the number of clusters that correspond to the sought-after states. The window segment size is a parameter that our domain expert collaborators are very confident setting up given their empirical knowledge stemming from working with the detector. The number of clusters can be relatively easily narrowed down by popular heuristics~\cite{schubert2023stop}, minimizing the time spent in hyperparameter tuning by the operators.

Our proposed algorithm for identifying different environmental states of the detector based on the subset of the channels we have selected is as follows:
\begin{enumerate}
    \item We are given a batch of $C$ time-series, corresponding to the different channels of interest. The current chosen length is one hour.
    \item For a given input window size $w$, create $C$ segments per non-overlapping window of the entire length.
    \item For each set of $C$ time-series corresponding to the same window, derive a set of statistical features. In this way, a given window becomes a data point.
    We experimented with (a)~simple statistics such as mean and standard deviation, (b)~\texttt{tsfresh}~\cite{christ2018time}, and (c)~\texttt{catch22}~\cite{lubba2019catch22}. In our use cases we found (a) to be sufficient, but if and when we introduce more complex patterns, more sophisticated features may become necessary.
    \item Run k-means clustering on all windows/data points. We utilized the \texttt{k-means++}~\cite{ilprints778} algorithm to find the initial seeds for the k-means clustering. We determine the number of clusters by intersecting the results with known intrinsic cluster validation indices~\cite{schubert2023stop}, see~\ref{fig:num_of_clusters}. Scikit-learn~\cite{JMLR:v12:pedregosa11a} library was used for all of the modeling done in this analysis.  
    \item As a final step, we need to provide a human-understandable set of labels for our results. Instead of using the raw names of the channels involved, which are not necessarily always intuitive, even to an experienced operator, we create three different ``replicas'' of the channels: (a) Anthropogenic, (b) Microseism, and (c) Earthquake. They correspond to known thresholds that LIGO operators currently manually use to identify one of those events. We then take the centroid of a given cluster and compare it against the known thresholds, and label the cluster accordingly.
\end{enumerate}
Figure~\ref{fig:pipeline} graphically illustrates this workflow.

\subsection{Downstream Evaluation}
\label{subsec:down_eval} 
Obtaining ground truth for our task is a rather ill-defined problem. For some of the discovered states, we can confirm the presence or absence of an earthquake by cross-referencing the discovered states with USGS data. However, less well-defined and properly monitored states, such as ones caused 

\begin{figure}[tbp]
    \begin{center}
        \includegraphics[width = 0.5\textwidth]{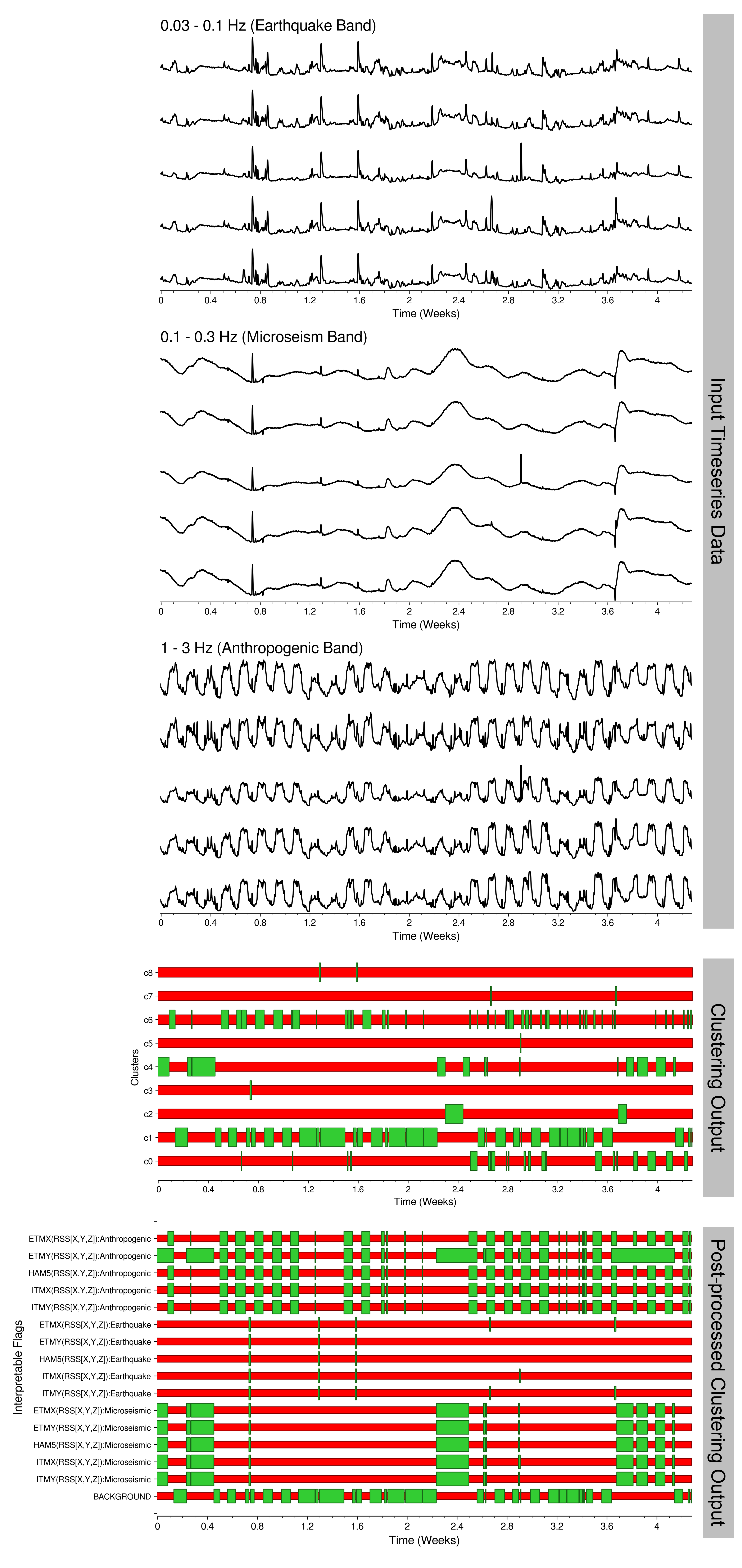}
        \caption{Example of our end-to-end analysis. Top panel consists of a set of sample channels that we are running the model on. The middle panel is the output of the clustering model. Each line represents a different state and the green flags are the segments at which the detector is in that specific state. The bottom panel shows the states assigned by the field experts using a simple threshold. It can be seen that the model's output recovers the expert's expectations without supervision. For example, there are clusters that correspond to earthquakes, high microseism, and high anthropogenic noise. }
        \label{fig:combined_result}
    \end{center}
\end{figure}

\noindent
by anthropogenic factors, are very hard to validate. Thus, in lieu of ground truth, we investigate if and to what degree the discovered states correlate with instances where it has been documented that the detector was witnessing a noise glitch or experienced a loss of lock.
\section{Results}
\label{results}
In this section, we first demonstrate an example of running our tool end-to-end and what we envision the final deployed outcome would look like. Then, we present results that link certain discovered states with documented problematic detector states.

\subsection{Indicative End-to-End Results}
\begin{figure}[ht]
    \begin{center}
        \includegraphics[width = 0.5\textwidth]{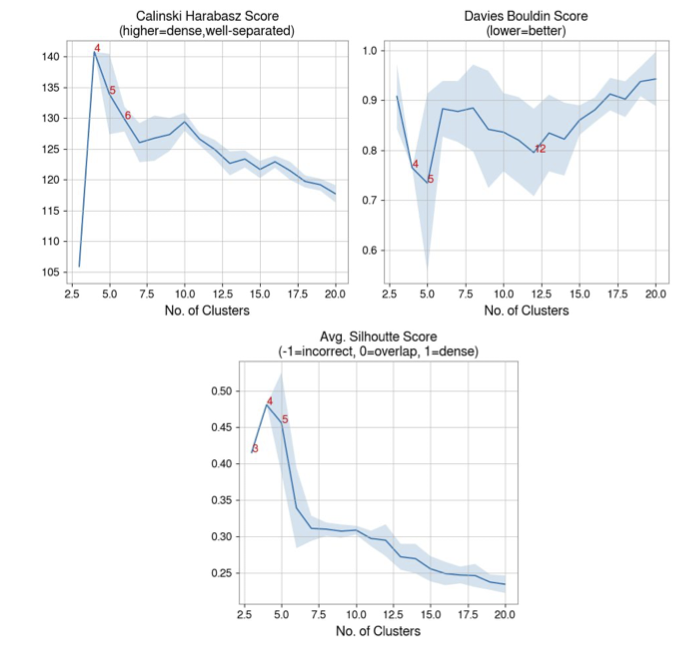}
        \caption{Cluster validation indices~\cite{schubert2023stop} employed in order to identify a short range of admissible number of cluster values that the operator can iterate over. A grid search for [3-20] number of clusters was done and three standard clustering validation scores where calculated. The final number of clusters was determined using the correlation with the Glitch rates as an external validation metric. }
        \label{fig:num_of_clusters}
    \end{center}
\end{figure}

Figure~\ref{fig:combined_result} shows a snapshot of our results, starting from the different channel time series as they are band-passed in different frequency bands (top), to the raw clustering results (center), to the labeled states (bottom). We envision the bottom panel of the figure to be the final product of the tool shown to the operator in near real-time. Figure~\ref{fig:num_of_clusters} shows an indicative set of results for identifying the number of clusters/states in a data sample. We find that different cluster validation indices have slight variation on the ``best'' number indicated. However, they seem to indicate a small range of admissible sets of states that are very feasible for the operator to iterate over and inspect.

\subsection{Linking Discovered States to Glitches and Loss of Lock}

\begin{figure*}[!t]
    \begin{center}
        \includegraphics[width = \textwidth]{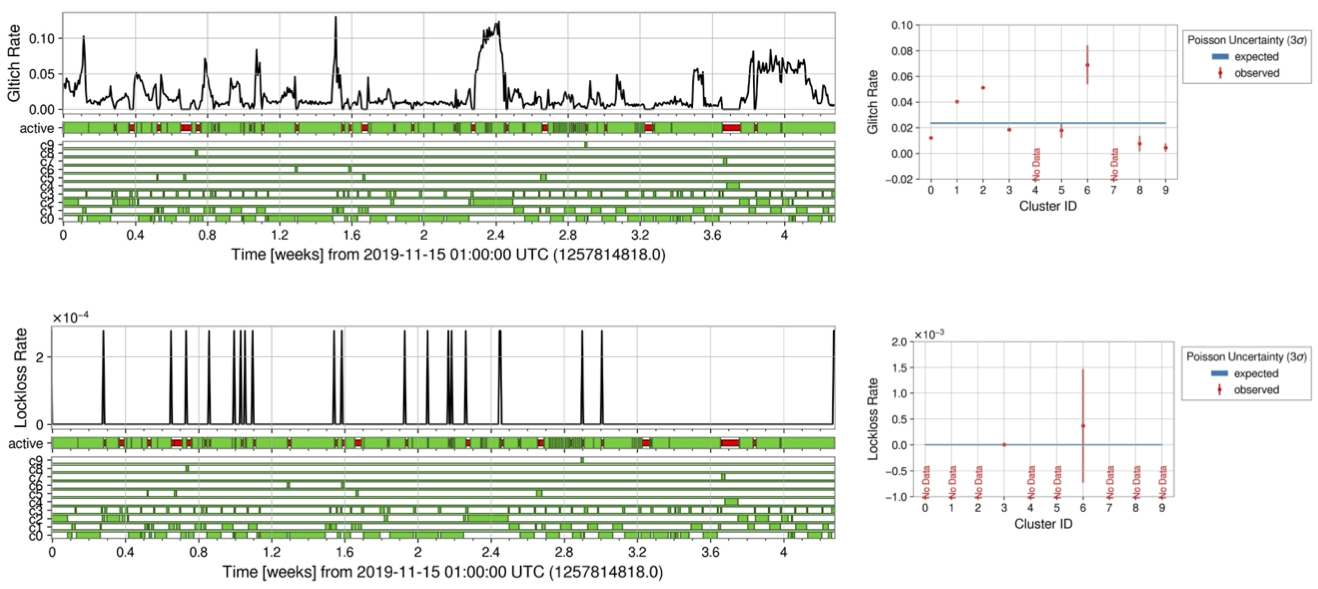}
        \caption{Some of our discovered states experience a far higher number of glitches than expected, assuming random occurrence (top). This indicates that there is a physical connection between the identified environmental states and those core detector problems, so our pipeline can provide valuable diagnostic information to guide future detector improvements and commissioning.}
        \label{fig:glitch_lockloss_correlation}
    \end{center}   
\end{figure*}

One of the main goals of our proposed monitoring tool is to be able to diagnose problematic detector states, such as periods of controls instabilities and elevated noise glitch activity~\cite{glanzer2023data}. In order to do so, we conducted an experiment where we computed the expected glitch and loss of lock rates, assuming they occur randomly (i.e., have no physical relation to the identified environmental states), and then compare that expected rate to the observed rate per discovered state. 
The observed glitch rate was calculated using the publicly available Gravity Spy glitch classifications dataset~\cite{glanzer2023data}, considering all the triggers with $SNR > 7.5$ during the analysis period. We make an extremely fascinating observation: For some states, the observed glitch rate far exceeds the expected rate, thus linking those states to those core detector issues. Figure~\ref{fig:glitch_lockloss_correlation} shows an example of such analysis, where some discovered states experience a much larger amount of glitches than would be randomly expected.
\section{Conclusion \& Future Work}
\label{conclusion}
In this paper, we present an instance of applied data science for detecting different environmental states of the LIGO detectors. This work has been directly motivated by working closely with LIGO commissioners and operators, understanding their needs, and translating them into a data science pipeline. We make interesting and important observations that link various discovered environmental states to documented issues faced by the detector, which is a positive step in the process of addressing those issues towards improving the detectors' up-time and the quality of the science data.

In the near future, we will continue working closely with LIGO for subsequent deployment of our tool. Furthermore, we are interested in extending our framework in order to be able to accommodate \textit{novel}, previously unrecognized environmental states. Finally, towards a highly-efficient deployment, we would like to explore the transition to a supervised model including known states discovered by our unsupervised tool and possibly obtain labeled data in a citizen science fashion, similar to the Gravity Spy project~\cite{glanzer2023data, zevin2024gravity} which does so for detecting glitches in the main channel of LIGO.
\section*{Acknowledgments}
This material is based upon work supported by NSF's LIGO Laboratory which is a major facility fully funded by the National Science Foundation. The authors are grateful for computational resources provided by the LIGO Laboratory and supported by the National Science Foundation under Award Nos. PHY-0757058 and PHY-0823459. Research at UC~Riverside was supported by the National Science Foundation under Award Nos. PHY-2141072 and IIS-2046086. This research has made use of data or software obtained from the Gravitational Wave Open Science Center (gwosc.org), a service of the LIGO Scientific Collaboration, the Virgo Collaboration, and KAGRA. This paper carries LIGO Document Number LIGO-P2400407.

\balance
\bibliographystyle{plain}
\bibliography{references}

\end{document}